# Artificial Intelligence-Based Opportunistic Coronary Calcium Screening in the Veterans Affairs National Healthcare System

Raffi Hagopian[1,2,3], Timothy Strebel[4], Simon Bernatz[9,10], Gregory A Myers[4], Erik Offerman[3], Eric Zuniga[3], Cy Y Kim[1,3], Angie T Ng[1,3], James A Iwaz[1,3], Sunny P Singh[2,4], Evan P Carey[2,4,5], Michael J Kim[2,4], R Spencer Schaefer[6], Jeannie Yu[1,3], Amilcare Gentili[7,8], Hugo JWL Aerts[9,10,11]

1. Division of Cardiology, Veterans Affairs Long Beach Healthcare System, Long Beach, CA, USA
2. Applied Innovations and Medical Informatics (AIMI), Veterans Affairs Long Beach Healthcare System, Long Beach, CA, USA
3. Division of Cardiology, University of California, Irvine, Irvine, CA, USA
4. Office of Research and Development, Veterans Health Administration, Washington DC, USA
5. Department of Biostatistics and Informatics, University of Colorado School of Public Health, Denver, CO, USA
6. Medical Informatics Data Science and Emerging Technology (MinDSET), Veterans Affairs Kansas City Healthcare System, Kansas City, MO, USA
7. Department of Radiology, Veterans Affairs San Diego Healthcare System, San Diego, CA, USA
8. Department of Radiology, University of California, San Diego, San Diego, CA, USA
9. Artificial Intelligence in Medicine (AIM) Program, Mass General Brigham, Harvard Medical School, Boston, MA, USA
10. Department of Radiation Oncology, Brigham and Women's Hospital and Dana-Farber Cancer Institute, Harvard Medical School, Boston, MA, USA
11. Radiology and Nuclear Medicine, GROW & CARIM Maastricht University, Maastricht, Netherlands

Correspondence: Raffi.Hagopian@va.gov

**Abstract**:

Coronary artery calcium (CAC) is highly predictive of cardiovascular events. While millions of chest CT scans are performed annually in the United States, CAC is not routinely quantified from scans done for non-cardiac purposes. A deep learning algorithm was developed using 446 expert segmentations to automatically quantify CAC on non-contrast, non-gated CT scans (AI-CAC). Our study differs from prior works as we leverage imaging data across the Veterans Affairs national healthcare system, from 98 medical centers, capturing extensive heterogeneity in imaging protocols, scanners, and patients. AI-CAC performance on non-gated scans was compared against clinical standard ECG-gated CAC scoring. Non-gated AI-CAC differentiated zero vs. non-zero and less than 100 vs. 100 or greater Agatston scores with accuracies of 89.4% (F1 0.93) and 87.3% (F1 0.89), respectively, in 795 patients with paired gated scans within a year of a non-gated CT scan. Non-gated AI-CAC was predictive of 10-year all-cause mortality (CAC 0 vs. >400 group: 25.4% vs. 60.2%, Cox HR 3.49, $p < 0.005$), and composite first-time stroke, MI, or death (CAC 0 vs. >400 group: 33.5% vs. 63.8%, Cox HR 3.00, $p < 0.005$). In a screening dataset of 8,052 patients with low-dose lung cancer-screening CTs (LDCT), 3,091/8,052 (38.4%) individuals had AI-CAC >400. Four cardiologists qualitatively reviewed LDCT images from a random sample of >400 AI-CAC patients and verified that 527/531 (99.2%) would benefit from lipid-lowering therapy. To the best of our knowledge, this is the first non-gated CT CAC algorithm developed across a national healthcare system, on multiple imaging protocols, without filtering intra-cardiac hardware, and compared against a strong gated CT reference. We report superior performance relative to previous CAC algorithms evaluated against paired gated scans that included patients with intra-cardiac hardware.

**Background**:

Atherosclerosis, the fundamental driver of cardiovascular disease, progresses over decades before patients present with clinical symptoms, myocardial infarction, stroke, or death. The natural history of atherosclerosis offers a unique opportunity for early detection and treatment, allowing a physician to slow the progression towards irreversible adverse events.[1] Traditional atherosclerotic cardiovascular disease (ASCVD) risk calculators use age, lipid levels, blood pressure, diabetes diagnosis, and smoking status to help physicians determine whether to place a patient on lipid-lowering medications for primary prevention. These calculators offer limited predictive power, and evolving evidence shows that imaging of plaque is a superior predictor of who would most benefit from the early initiation of these therapies.[2]

Coronary artery calcium quantification on ECG-gated CT scans by the Agatston formula is a well-established method of determining atherosclerotic cardiovascular risk.[3] Since its proposal in the 1990s, Agatston scoring of coronary calcium has demonstrated utility in predicting all-cause mortality and major adverse cardiovascular events (MACE). This has been confirmed in longitudinally followed patient cohorts, such as the Multi-Ethnic Study of Atherosclerosis (MESA).[4] Recent AHA/ACC guidelines recommend the use of coronary calcium scoring to help reclassify patients with borderline cardiovascular risk, and the use of cardiac CT is an evolving area in medical practice as scanner technology improves and interpretation skills disseminate.[5] While ECG-gated CTs remain the gold standard for quantifying coronary artery calcium, most chest CT scans obtained for routine clinical purposes are not ECG-gated, meaning they are not electrocardiographically synchronized to heartbeat to reduce cardiac motion during the scan. While non-ECG gated CT scans have relatively poor cardiac anatomy definition due to cardiac motion, physicians can often review these studies qualitatively to identify coronary calcium, which may inform the decision to start a patient on lipid-lowering therapies. Prior studies have shown high correlation between manually derived CAC scores from non-gated studies compared to gated studies, and a meta-analysis reported a pooled correlation coefficient of 0.94 (95% CI 0.89-0.97).[6]

Various research groups have investigated the potential application of deep learning algorithms towards the automation of coronary calcium scoring on non-gated ECG studies.[7,8,9,10,11,12,13] Coronary calcium has a distinctive visual appearance on CT and is labeled with low inter-reader variability, making it a particularly tractable problem for modern deep learning segmentation algorithms. Zelesnik et al. showed that these automated CAC scores from non-gated studies predict outcomes across a variety of large symptomatic and asymptomatic patient cohorts.[9] Eng et al. expanded on this problem by benchmarking their non-ECG gated AI predictions against paired ECG-gated CTs and exploring an end-to-end neural network architecture that directly segments coronary calcification without first requiring cardiac segmentation.[10] More recently, initial prospective testing of these models in clinical settings has begun, but these systems are not disseminated in widespread clinical practice and have yet to become part of routine clinical care.[14] Limitations in prior studies using AI to quantify CAC from non-gated CT scans include: not testing against a strong gated CT CAC-scoring standard, testing on a single CT protocol, testing on a limited number of medical centers, and not being trained or evaluated on patients with intracardiac hardware.

Building upon previous AI coronary calcium scoring studies, we developed and evaluated our own AI model within the Veterans Affairs national healthcare system, using data from 98 distinct medical centers across the United States. The VA was one of the first healthcare systems to adopt electronic medical records systems in the 1990s and has excellent patient retention longitudinally because it serves as a single-payer healthcare network for 14 million veterans. This results in an ideal dataset for developing predictive models, as long-term outcome, diagnosis, and medication information are available across an extensive network of hospitals. We use a diverse set of non-gated CT imaging protocols as input for our AI-CAC model and compare its predicted CAC scores against the reference standard of gated-CT CAC scoring. Additionally, we did not filter out intracardiac hardware, which makes CAC quantification more challenging. Relative to prior works, we chose to create challenging datasets to develop a model that would be more robust to real-world clinical variation.

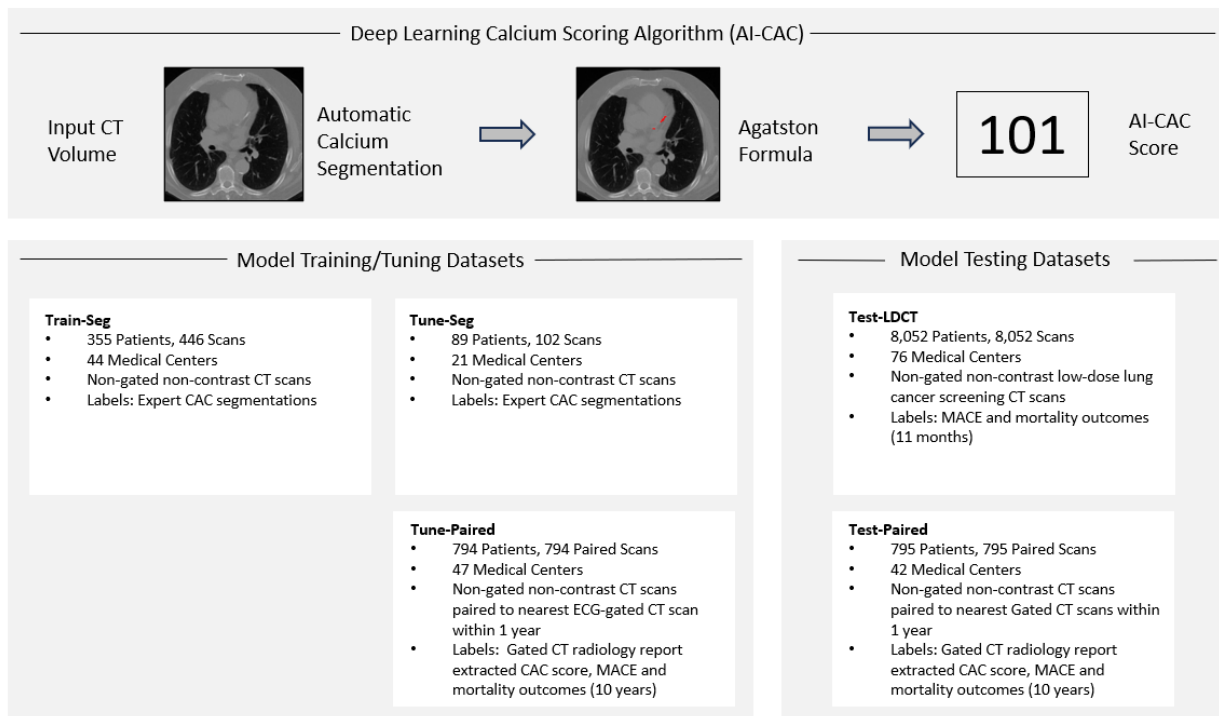

**Figure 1. Algorithm:** Non-gated CT scans are processed using a trained U-Net-variant neural network model that produces coronary artery calcium segmentation masks. These masks along with CT voxel Hounsfield units are entered into the Agatston formula to compute a CT scan-level score. **Datasets:** Our model was trained on 446 non-gated scans with manual segmentations ("Train-Seg") and tuned on 102 non-gated scans with manual segmentations ("Tune-Seg"). Datasets consisting of paired gated- and non-gated CTs within a year of each other from the same patient were created to compare non-gated scan AI-CAC scores to gated radiology report Agatston scores for hyperparameter tuning ("Tune-Paired") and model testing ("Test-Paired"). A dataset of 8,052 low-dose lung cancer screening scans was collected to simulate opportunistic screening ("Test-LDCT"). This dataset had longitudinal outcomes available, but no reference gated CT report CAC scores or manual CAC segmentations. There were no shared patients between datasets. Dataset characteristics are further detailed in **Supplemental Table 1**.

**Methods**:
This research was Health Insurance Portability and Accountability Act compliant and approved by local Institutional Review Boards and Research and Development Committees. A waiver of informed consent was obtained. The Long Beach VA Healthcare System served as the primary research site.

*Dataset Construction (Segmentation and Paired Datasets):*

Patients who had undergone a non-contrast, non-gated thoracic CT within a year of a gated CT scan for CAC scoring were selected from internal VA databases using CPT codes 71250, 71270, and 75571. This cohort construction methodology was similar to Eng et al. but we used a narrower time window between our gated- and non-gated CT pairs to further limit coronary calcium progression between scans.[10,15] Only scans obtained prior to January 1st, 2020 were selected to increase follow-up duration and avoid any potential impact of the COVID-19 pandemic on model development. Patient clinical outcomes were collected to an end-point of May 1st, 2024. This process identified 10,265 paired CT studies (4,179 gated, 6,086 non-gated) from 4,063 distinct patients. Due to limitations in mapping from internal SQL databases to the VA's internal VistA Imaging platform, the retrieval process yielded only 8,499 of the requested paired studies. Of the retrieved studies, 7,182 scans from 3,595 patients passed quality filtering removing non-chest anatomy, contrast, post-processed studies, non-axial series, and filtering to slice thicknesses of 2.5 mm to 5 mm. Out of 4,179 gated reports, 3,397 had an extractable Agatston score. DICOM metadata filtering and NLP techniques for Agatston score extraction are further described in **Supplementary Materials**. After accounting for issues with imaging data retrieval, imaging quality control filtering, and extraction of CAC scores from gated CT radiology reports, these patients were divided into distinct datasets **Figure 1**. The "Tune-Paired" and "Test-Paired" datasets were created from 1,589 distinct patients with a complete set of paired non-gated and gated scans, extractable gated report CAC scores, and unambiguous mapping of images to reports. These 1,589 patients underwent a stratified-random 50:50 split by medical center to balance the representation of scanners/protocols between the tuning and testing paired datasets. Coronary calcium was manually segmented by a cardiac CT board-certified cardiologist in Slicer3D in 548 non-gated CT studies from 444 patients.[16] While coronary calcium is not conventionally annotated on non-gated scans, a similar manual process to conventional scoring was followed by identifying voxels above 130 Hounsfield Units. The 444 patients with expert non-gated calcium annotations underwent an 80:20 split to create the "Train-Seg" and "Tune-Seg" datasets, used for model training and tuning respectively. There were no shared patients between datasets.

*Dataset Construction (LDCT Dataset):*

To simulate prospective screening, a dataset of non-gated low-dose lung cancer-screening CTs (LDCT) with CPT code 71271 were retrieved in reverse chronological order from the present. The "Test-LDCT" cohort was not limited to scans prior to the COVID-19 pandemic to facilitate understanding of model performance on contemporary scans. A convenience sample of 10,425 LDCT studies was retrieved of which 8,057 studies (8,052 unique patients) passed previously described quality controls and exclusion of patients in other training, tuning, or testing datasets. Only the oldest scan was kept for the five patients with more than one LDCT scan, resulting in 8,052 total studies.

*Model Development:*

We developed a deep learning model that directly segments coronary calcium on non-contrast enhanced, non-gated CT scan axial slices. Manual CAC labels on 446 non-gated scans were used to train a 2D Swin-UNETR model, a modern transformer architecture-based variant of the U-Net that is well-suited for anatomic segmentation tasks on CT.[17,18] Our model was trained on a single AWS node with 4 GPUs for 35 epochs. We used an initial learning rate of 1e-3 with stepwise rate decay, a dropout rate of 0.2, weight decay of 1e-5, and random affine transformations for data augmentation. Consistent with Eng et al. we found that a training curriculum starting with only slices that contained coronary calcium then subsequently introducing slices without coronary calcification that the model erroneously labels as containing calcium was an efficient training strategy.[10] In addition, we found using FocalLoss, a variant of cross entropy, worked well for coronary calcium segmentation consistent with this function's utility in amplifying sparse signals. The predicted masks were processed using connected-component analysis to identify contiguous lesions, and the Agatston formula was applied to obtain a CT scan-level AI-CAC score **Figure 1**.

*Model Evaluation:*

Univariate Kaplan-Meier curves were constructed using AI-CAC scores (binned into 0, 1-100, 101-400, and >400) for all-cause mortality, and composite events (first MI, stroke, or all-cause mortality) in the "Test-Paired" and "Test-LDCT" datasets. Follow-up duration up to 10 years is shown for the "Test-Paired" dataset and up to 11 months for the "Test-LDCT" dataset. The latter dataset had shorter follow-up duration as scans were obtained closer to the present to assess model performance on contemporary scans. For the "Test-LDCT" dataset we also show outcomes separated by whether a patient was ever prescribed lipid-lowering therapies in their lifetime (for all-cause mortality) or prior to their first event (for composite MI, CVA, and death). Additionally, Kaplan-Meier curves were constructed using the "Test-LDCT" dataset after excluding medical centers in the "Train-Seg" dataset to assess model generalization to hospitals that were not used for segmentation model training ("Exclusive-Test-LDCT").

Confusion matrices and log-transformed scatter plots were constructed for non-gated AI-CAC scores compared to expert non-gated CAC scores on the "Tune-Seg" dataset. Similar figures were created for non-gated AI-CAC scores compared to paired conventional Agatston scores from gated CT radiology reports in the "Tune-Paired" and "Test-Paired" datasets. In addition, expert manual non-gated CT CAC scores from the "Train-Seg" and "Tune-Seg" datasets were compared against gated CT report CAC scores, for cases where gated CT pairs were available. Correlation coefficients (Spearman, Pearson, and ICC) are reported for the dot plots prior to log-transformation and linearly-weighted Kappa statistics are reported for the confusion matrices. For further details on statistical analysis see **Supplementary Materials.**

AI-CAC scores on "Test-LDCT" studies with AI-CAC > 400 were randomly sampled for cardiologist review. For each of the randomly selected studies with elevated AI-CAC, slices that were predicted to contain coronary calcium and the model's calcium masks were saved and shown to one of four cardiologists. The cardiologists were asked to review the positive predicted slices for each CT study and qualitatively determine if there was sufficient calcification to recommend initiation of lipid-lowering therapies (classified as 'correct'), borderline calcification on the positive slices ('uncertain'), or minimal to no coronary calcification ('incorrect').

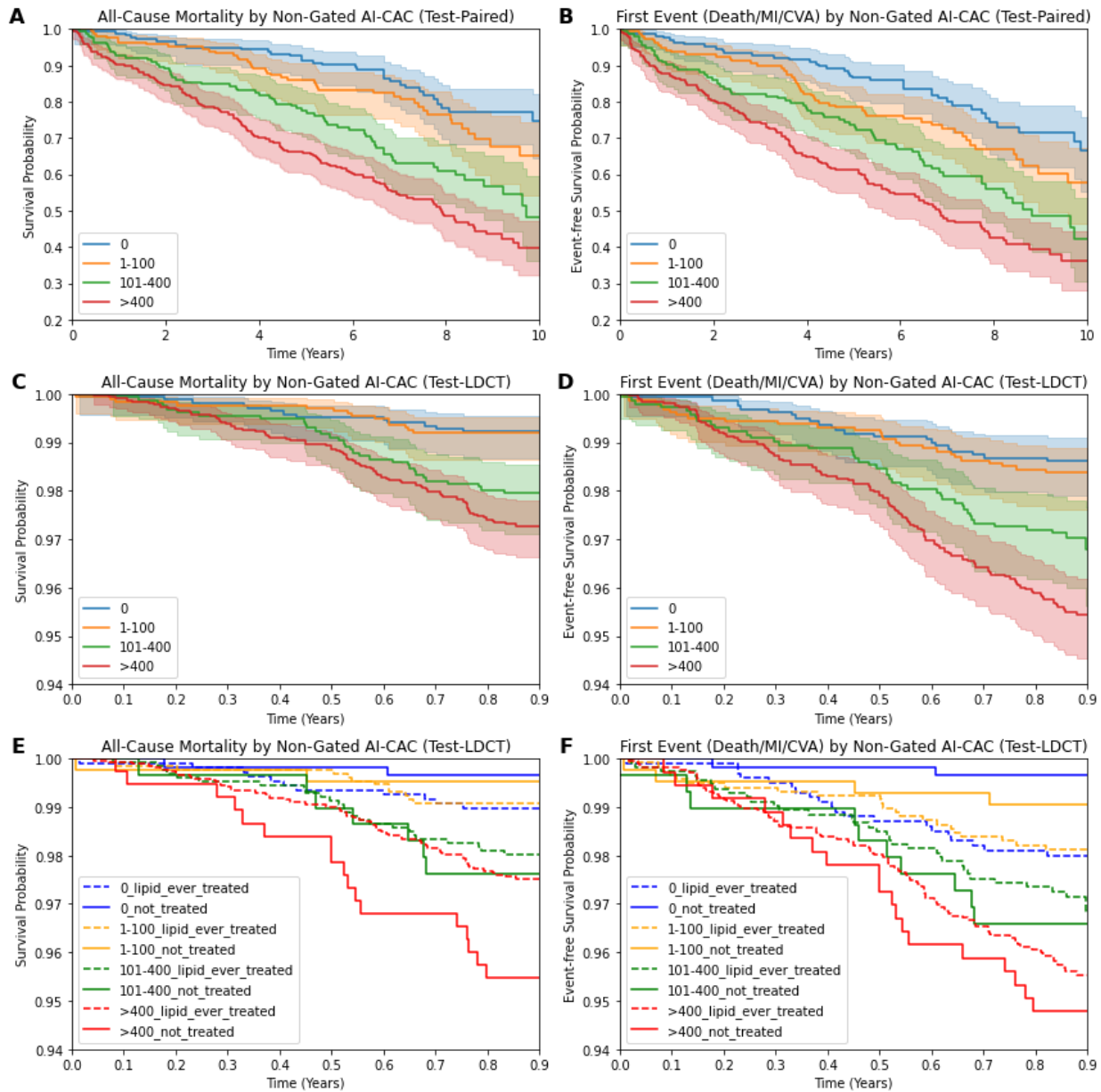

Figure 2. **Univariate Kaplan-Meier Curves.** All-cause mortality curves are shown on the left **(A, C, E)**, and composite initial occurrence of MI/Stroke/Death is shown on right **(B, D, F).** For the composite event curves, individuals who had a MI or stroke event before their non-gated CT scan were filtered out. Differences in axis scales for "Test-Paired" **(A, B)** and "Test-LDCT" **(C, D, E, F)** datasets are due to different follow-up durations. Plots **(E, F)** separate patients by whether they have ever been prescribed lipid-lowering therapies within their lifetime **(E)** or prior to their first composite event **(F)** regardless of compliance.

**Results**:

AI-CAC on non-gated scans was predictive of 10-year all-cause mortality (CAC 0 vs. >400: 25.4% vs. 60.2%, Cox HR 3.49, p < 0.005), and the composite of initial stroke, MI, or death (CAC 0 vs. >400: 33.5% vs. 63.8%, Cox HR 3.00, p < 0.005) on the "Test-Paired" dataset **Figure 2A, B**. Clinical standard CAC scores extracted from gated CT reports demonstrated comparable stratification of all-cause mortality over 10 years (CAC 0 vs. >400: 22.8% vs.

53.4%, Cox HR 3.17, p < 0.005) **Supplemental Figure 1**. In the opportunistic screening "Test-LDCT" dataset of 8,052 individuals, significant differences were observed by AI-CAC group at 11 months of follow-up for all-cause mortality (CAC 0 vs. >400: 0.8% vs. 2.7%, Cox HR 3.53, p < 0.005) and for composite events (CAC 0 vs. >400: 1.4% vs. 4.6%, Cox HR 3.22, p < 0.005) **Figure 2C, D**. Further stratification of "Test-LDCT" patients with AI-CAC >400, based on whether they had ever been prescribed lipid-lowering therapies (regardless of compliance), showed a trend toward mortality reduction (2.5% vs. 4.5%, ARR = 2.0%), but did not reach statistical significance by 11 months of follow-up **Figure 2E**.

Filtering the "Test-LDCT" dataset (76 medical centers) to exclude scans from the 44 centers represented in the training "Train-Seg" dataset, resulted in the "Exclusive-Test-LDCT" dataset (3,931 patients; 48 medical centers). At 11 months of follow-up, significant differences in all-cause mortality (CAC 0 vs. >400: 0.9% vs. 2.8%, Cox HR 3.11, p = 0.01) and composite events (CAC 0 vs. >400: 1.4% vs. 4.8%, Cox HR 3.53, p < 0.005) remained in the "Exclusive-Test-LDCT" dataset suggesting generalization to hospital systems that were not used for model training **Supplemental Figure 2**. AI-CAC outcome by dataset is summarized in **Supplemental Table 2**.

AI model performance was compared to expert performance on non-conventional calcium scoring of non-gated CT chest studies on the "Tune-Seg" dataset of 102 studies manually segmented by a cardiac CT board-certified cardiologist. These segmentations were not used for model training. AI-CAC scores were highly correlated with expert non-gated scan CAC scores (ICC = 0.96, Spearman r = 0.90) and had high agreement across CAC groups of 0, 1-100, 101-400, and >400 (Kappa 0.81) **Figure 3A and Supplemental Figure 3A.**

To determine how non-gated AI-CAC scoring performs against the clinical standard of radiologist scoring of gated CT scans, we compared non-gated AI-CAC scores to CAC scores extracted from gated CT radiology reports. The "Test-Paired" dataset consists of 795 patients who had a paired non-gated CT within one year of a gated CT scan (mean [±std]: 4.3 ± 3.3 months). The "Test-Paired" dataset was not used for model training or hyperparameter tuning. AI-CAC had an accuracy of 89.4% (F1 0.93) at differentiating zero vs. non-zero CAC, and 87.3% accuracy (F1 0.89) at differentiating less than 100 vs. 100 or greater CAC. Performance of non-gated AI-CAC relative to gated CAC scores at various dichotomous thresholds is shown in **Table 1**. AI-CAC vs. gated report CAC showed good agreement between groups of 0, 1-100, 101-400, >400 ("Test-Paired": Kappa = 0.72) **Figure 3.** Subgroup analysis was also performed to assess the performance of the model across demographics of sex, scanner manufacturer, and KVP in our tuning and testing datasets **Supplemental Table 3**. The model showed strong performance across sex despite women consisting of only 5.9% of our training dataset.

After development, we ran the model on 8,052 patients with non-gated low-dose lung cancer-screening CTs (LDCT) to simulate opportunistic screening. In this dataset, 20.7%, 21.4%, 19.4%, and 38.4% had non-gated AI-CAC scores of 0, 1-100, 101-400, and >400, respectively. Of the living "Test-LDCT" dataset patients with AI-CAC >400, 30.6% (920/3,007) were not on lipid-lowering therapies at the end of the follow-up period. Out of 531 randomly selected "Test-LDCT" patients with AI-CAC >400, cardiologists' qualitative review determined that 527/531 (99.2%) would benefit from lipid-lowering therapy **Supplemental Figures 8, 9.** Representative images of model segmentation masks on the "Test-LDCT" dataset are shown in the **Supplementary Materials.**

| CAC Threshold | Accuracy(%) | PPV(%) | NPV(%) | Sensitivity(%) | Specificity(%) | F1(%) |
|---|---|---|---|---|---|---|
| 1 | 89.4 | 94.8 | 70.5 | 91.9 | 79.5 | 93.3 |
| 100 | 87.3 | 91.9 | 81.2 | 86.7 | 88.2 | 89.2 |
| 400 | 87.4 | 86.9 | 87.7 | 80.2 | 92.1 | 83.4 |

**Table 1. Non-Gated AI-CAC Performance Compared to Gated CT Report CAC by Dichotomous CAC Score Threshold.** This table shows statistics calculated on the "Test-Paired" dataset. It depicts whether non-gated AI-CAC versus gated-report CAC scores are congruent with respect to the listed dichotomous thresholds (equal to threshold or greater vs. below threshold). These shown thresholds are often used by clinicians for the decision of whether to initiate lipid-lowering therapies or to evaluate risk.

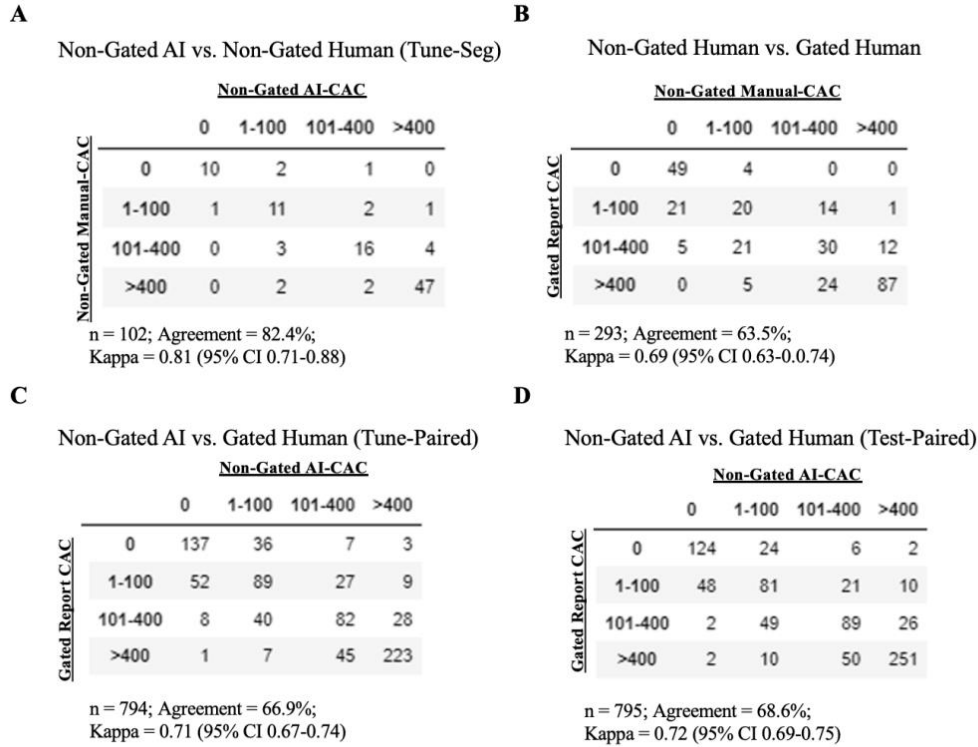

**Figure 3. Confusion matrices. (A)** Non-gated AI-CAC vs. non-gated manual segmentation on the "Tune-Seg" dataset. **(B)** Non-gated manual segmentation compared to paired gated CT report scores on a combination of "Train-Seg" and "Tune-Seg" datasets studies that had paired scans available within a year. **(C)** Non-gated AI-CAC vs. paired gated CT report scores on "Tune-Paired" dataset and **(D)** against the "Test-Paired" dataset. Linearly-weighted Cohen's Kappa with bootstrapped 95% confidence intervals and percent agreement are shown.

**Discussion:**

This study demonstrates that non-gated scan AI-CAC scores can predict long-term outcomes across the Veterans Affairs system in both symptomatic patients (paired gated/non-gated) and asymptomatic patients (LDCT) despite the short follow-up duration in the latter dataset. Furthermore, this predictive ability generalized to 48 medical centers not present in the training dataset. At present, VA imaging storage systems contain millions of non-gated, non-contrast chest CTs, but less than fifty thousand gated studies. This asymmetry represents an opportunity for AI-CAC to leverage routinely collected non-gated scans for purposes of cardiovascular risk evaluation.[14] While cardiologists and primary care physicians already review non-gated scans for the presence of CAC, an AI solution can be scaled over the entire national enterprise. This would enable the rapid identification of patients at the highest cardiovascular risk and facilitate timely initiation of therapies to prevent irreversible events.

  We introduce a deep learning segmentation model that achieves near expert-level accuracy in quantifying coronary artery calcium from non-contrast, non-gated chest CT scans. The model generates segmentation masks that serve as a human-verifiable output, allowing physicians to rapidly confirm the presence of coronary calcification. Given the distinctive high-attenuation appearance of coronary calcium on CT, it is easier and faster to verify than many other imaging findings. Beyond identifying high-risk patients, AI-CAC can be used to filter slices within a patient's scan, presenting physicians only slices predicted to contain CAC to expedite the verification of patients who screen positive.

  Compared to prior works, our approach is different in many respects. In addition to drawing data from a heterogenous national cohort, we did not filter out patients with pacemakers, mechanical valves, coronary stents, or prior cardiac surgeries as previous studies have done.[21,22,23] Although these characteristics can be filtered out using electronic medical record data, we believe it is more prudent to train a model which avoids pacemakers and prostheses as an added precaution in the event that patient chart data is inaccurate or missing. While direct performance comparisons are difficult between published studies due to dataset differences, we report superior non-gated AI performance Kappa and F1 statistics when compared to works that used a similar stringent paired ECG-gated reference standard and did not explicitly filter out intracardiac hardware [Pieszko: Linearly-weighted Kappa 0.62 (95% CI 0.6-0.64) vs. 0.72 (95% CI 0.69-0.75) (ours); Eng: F1 at a 100 score threshold 0.458 vs. 0.892 (ours), and zero threshold 0.853 vs. 0.933 (ours)].[10,11]

Despite the tremendous potential of artificial intelligence to revolutionize the interpretation of medical images, several challenges remain preventing their widespread use in medicine. One of the main barriers of AI model adoption is that models often can learn features that are specific to a given training set, medical center, scanner, or protocol type and fail to generalize to a real-world setting. By training and evaluating a deep learning model across many medical centers with heterogeneous scanners and imaging protocols, we show that it is possible to develop a robust AI system that can work in a high-variability real-world setting across a national network of distinct medical centers.

While we aimed to simulate prospective testing as closely as possible with our LDCT dataset, ultimately the real test of a model's utility will be its evaluation in a prospective clinical environment. A potential limitation of CAC algorithms is their robustness to internal hardware. Unlike previous works, we did not filter out intrathoracic hardware; however, rare or new intrathoracic devices that the model has never been trained upon may affect model performance in unexpected ways. Periodic evaluation will be necessary to identify future issues. Additionally, while our dataset was large, establishing even larger datasets will allow us to capture greater patient-level variation and potentially improve performance.

In the future, we aim to achieve system-wide implementation of automated CAC scoring to improve medical outcomes in the veteran population that carries a high burden of cardiovascular disease. This model can be used to both prioritize patients at highest ASCVD risk (those with the highest CAC scores), and detect patients that had previously unknown atherosclerotic disease resulting in earlier or more aggressive lipid-lower therapy initiation. The sooner a patient's atherosclerotic plaque is detected and treated, the greater the benefit, as cardiovascular disease worsens proportionally to cholesterol exposure over time.[24] We believe utilizing medical imaging AI for use cases such as coronary calcium detection can help shift modern medicine from a reactive approach to the proactive prevention of disease, resulting in long-term reductions in morbidity, mortality, and overall healthcare costs.

**Acknowledgements**: We would like to thank Kenneth J. Graham and Brandon B. Konkel for their technical expertise in standing up imaging acquisition and cloud compute infrastructure for the VA, Thomas Gilcrest and Ning Xie for their administrative support dealing with evolving institutional AI policy, Dr. Morton J. Kern for editing the manuscript, and Dr. Peter H. Nguyen for his expertise in VA databases.

**Supplementary Materials:**

*Natural Language Processing to Extract Agatston Scores from Gated CT Radiology Reports:*

Agatston scores from the conventional CAC-scoring gated CT radiology reports were extracted using traditional natural language processing (NLP) techniques such as regular expressions. Out of 4,179 gated reports, 3,397 had an extractable Agatston score. Radiology reports without an extractable score often mentioned that scoring was not performed due to the presence of a stent or coronary artery bypass surgery. Three of the authors manually verified the NLP extracted Agatston scores from radiology reports in a random sample of 300 reports (100 per reviewer) and found that our script accurately extracted 99% of calcium scores in our random sample.

*Database Extraction of Clinical Outcomes and Medication Information:*

Clinical data for patients in our imaging datasets was retrieved by querying the internal VA databases. For each patient, we obtained mortality information including dates of death, myocardial and cerebral infarction diagnoses using ICD codes with corresponding entry dates, and lipid-lowering medication prescriptions with initial issue dates. Myocardial infarction was identified using ICD-10 code "I21" and cerebral infarction using "I63". Cross-referenced ICD-9 codes corresponding to these ICD-10 codes were also retrieved.

*DICOM Metadata Filtering to Select Optimal Series per CT Study:*

As each CT scan consists of multiple image series, DICOM metadata was programmatically filtered to select one series per study. Only series in axial orientation, not containing contrast, and with a slice thickness of 2.5 to 5 mm were kept to retain images similar to those used for conventional CAC scoring, which is performed on 3 mm slices. Additionally, we preferentially selected "Series Description" tags that contained the words cardiac, calcium, or lung. If a specific series protocol was repeated within a study, only the series with the latest timestamp was kept, as technicians may reacquire an imaging series if there were issues with the initial acquisition.

*Statistical Analysis:*

We report Pearson, Spearman, and Intraclass correlation (two-way mixed, single measurement, agreement) coefficients on our AI-derived CAC scores on non-gated images vs. expert-derived CAC on the non-gated images and on our AI-derived CAC scores on non-gated studies with paired expert-derived CAC on gated conventional CAC-scoring CTs. Accuracy, sensitivity, specificity, positive predictive values, negative predictive values, and F1 statistics were calculated for various calcium score dichotomous thresholds. Cohen's linearly-weighted Kappa statistics are shown for confusion matrices with CAC groups of (0, 1-100, 101-400, >400). Bootstrapping with 1,000 iterations was used to estimate 95% confidence intervals for linear-weighted Kappa statistics. For Kaplan-Meier graphs, univariate Cox-proportional hazard ratios (Cox HR) are reported with corresponding p-values.

| Dataset | Train-Seg | Tune-Seg | Tune-Paired | Test-Paired | Test-LDCT |
|---|---|---|---|---|---|
| Number of Patients | 355 | 89 | 794 | 795 | 8052 |
| Number of Women | 21 (5.9%) | 3 (3.4%) | 75 (9.4%) | 46 (5.8%) | 642 (8.0%) |
| Mean (Std Dev.) age in years at NG scan | 66.9 (9.6) | 69.1 (9.3) | 65.0 (9.5) | 65.9 (9.2) | 66.7 (7.5) |
| Study Date Range | 2010-2019 | 2010-2019 | 2009-2019 | 2009-2019 | 2023-2023 |
| Number of Medical Centers | 44 | 21 | 47 | 42 | 76 |
| Number of Non-Gated Studies | 446 | 102 | 794 | 795 | 8052 |
| Number of Gated Studies | NA | NA | 794 | 795 | NA |
| Toshiba/Canon NG Scans (%) | 201 (45.1%) | 41 (40.2%) | 363 (45.7%) | 373 (46.9%) | 2397 (29.8%) |
| GE NG Scans (%) | 131 (29.4%) | 31 (30.4%) | 284 (35.8%) | 269 (33.8%) | 3160 (39.2%) |
| Philips NG Scans (%) | 66 (14.8%) | 18 (17.6%) | 76 (9.6%) | 80 (10.1%) | 1165 (14.5%) |
| Siemens NG Scans (%) | 46 (10.3%) | 12 (11.8%) | 71 (8.9%) | 73 (9.2%) | 1330 (16.5%) |
| Hitachi NG Scans (%) | 2 (0.4%) | 0 (0.0%) | 0 (0.0%) | 0 (0.0%) | 0 (0.0%) |
| CAC 0 Gated Reports (%) | NA | NA | 183 (23.0%) | 156 (19.6%) | NA |
| CAC 1-100 Gated Reports (%) | NA | NA | 177 (22.3%) | 160 (20.1%) | NA |
| CAC 101-400 Gated Reports (%) | NA | NA | 158 (19.9%) | 166 (20.9%) | NA |
| CAC >400 Gated Reports (%) | NA | NA | 276 (34.8%) | 313 (39.4%) | NA |

**Supplemental Table 1. Dataset Characteristics.** Each column corresponds to a different dataset used for model training, tuning, or testing. There was no overlap of patients across datasets. For the "Paired" datasets, each patient had a non-gated and gated CT scan pair within a year time window. We kept the temporally nearest pair if a patient had more than one non-gated/gated pair (mean 4.3 months). Segmentation ("Seg") sets correspond to non-gated studies that had an expert manual coronary calcium segmentation mask for training or tuning purposes.

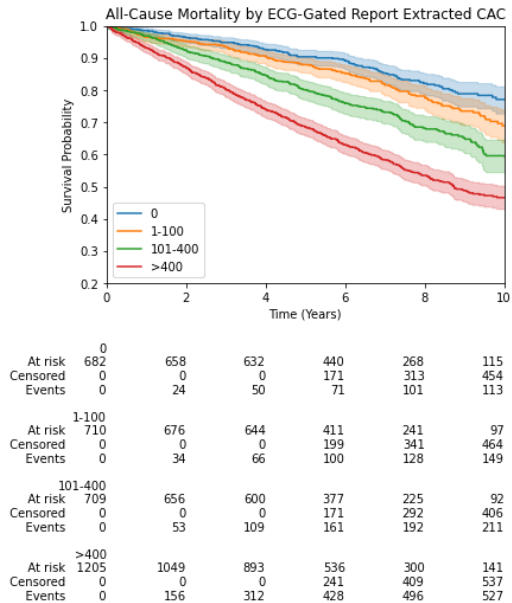

**Supplemental Figure 1.** All-cause mortality stratified by gated CT report extracted clinical CAC scores. If multiple gated scans were available for a given patient, the oldest report was used so that events were counted once per patient.

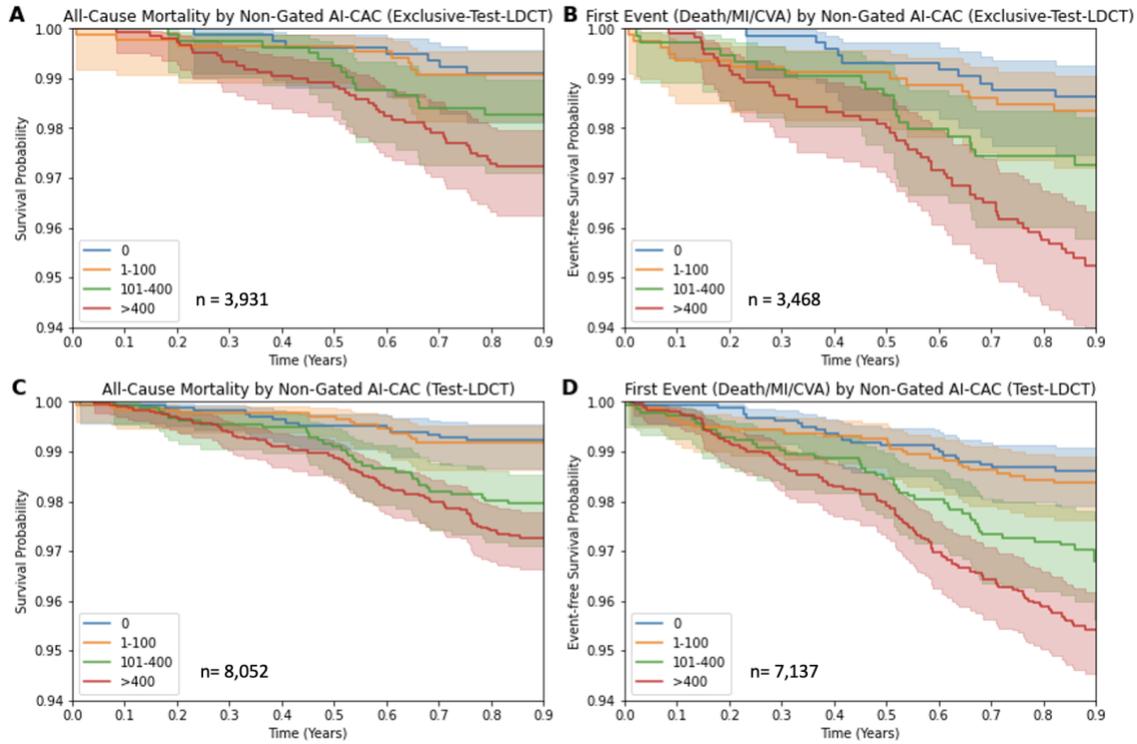

**Supplemental Figure 2.** Univariate Kaplan-Meier Curves for the "Test-LDCT" and "Exclusive-Test-LDCT" datasets that consisted of data from 76 and 48 medical centers, respectively. The "Exclusive-Test-LDCT" dataset was a subset of "Test-LDCT" after filtering out studies from any of the 44 medical centers that were represented in the "Train-Seg" dataset. These curves show the model's predictions generalized to medical centers not represented in the training dataset.

| Dataset | Follow-up Duration | Outcome | Event Rate CAC = 0 | Event Rate CAC > 400 | Cox Proportional Hazard Ratio | p-value |
|---|---|---|---|---|---|---|
| **Test-Paired** | 10 years | All-Cause Mortality | 25.4% | 60.2% | 3.49 | <0.005 |
| **Test-Paired** | 10 years | Initial MI, CVA, or Death | 33.5% | 63.8% | 3.00 | <0.005 |
| **Test-LDCT** | 0.9 years | All-Cause Mortality | 0.8% | 2.7% | 3.53 | <0.005 |
| **Test-LDCT** | 0.9 years | Initial MI, CVA, or Death | 1.4% | 4.6% | 3.22 | <0.005 |
| **Exclusive-Test-LDCT** | 0.9 years | All-Cause Mortality | 0.9% | 2.8% | 3.11 | 0.01 |
| **Exclusive-Test-LDCT** | 0.9 years | Initial MI, CVA, or Death | 1.4% | 4.8% | 3.53 | <0.005 |
| **Gated CT Human Scores** | 10 years | All-Cause Mortality | 22.8% | 53.4% | 3.17 | <0.005 |

**Supplemental Table 2.** Outcome comparison of non-gated scan AI-CAC of 0 vs. > 400 across the testing datasets. "Exclusive-Test-LDCT" is defined in the prior figure. The bottom row shows 10-year all-cause mortality for clinical CAC scores from gated scan reports.

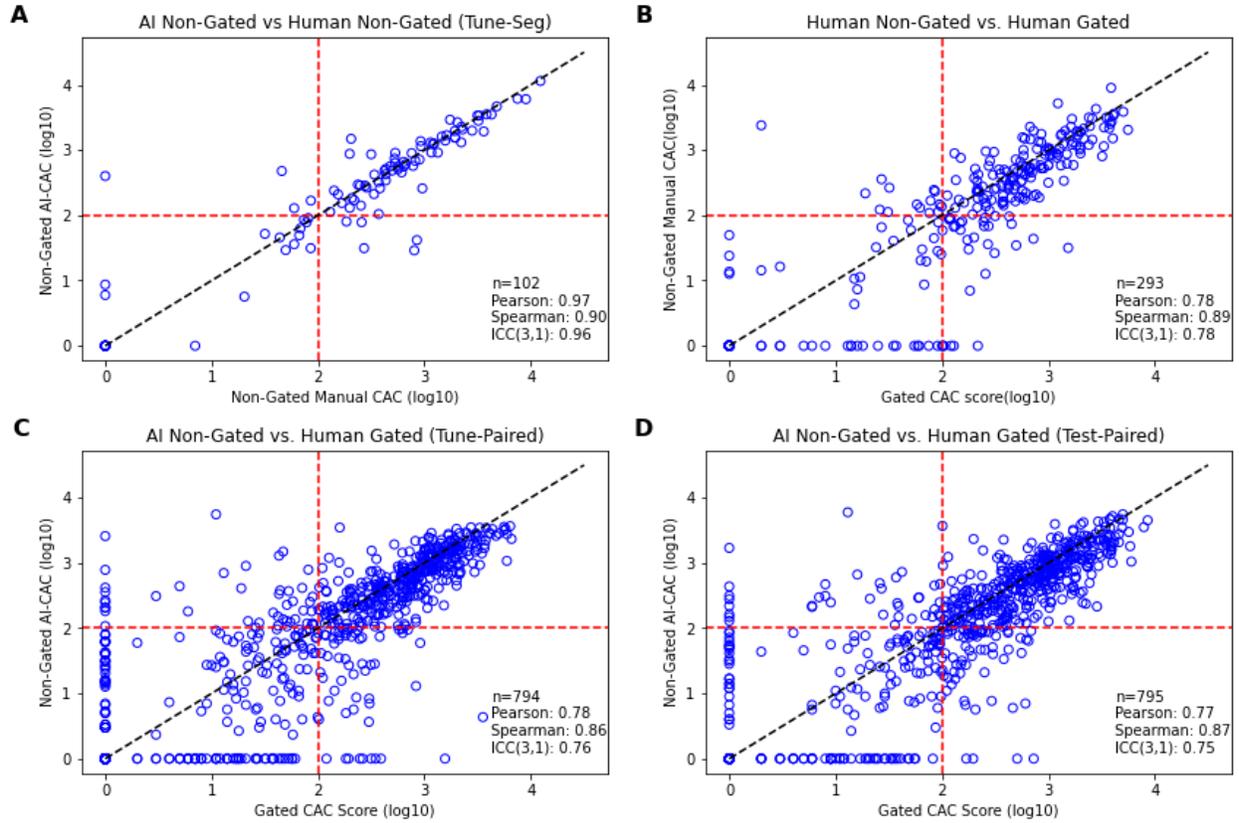

**Supplemental Figure 3. Dot Plots Comparing Modality, AI, and Human.** (**A**) Comparison of non-gated AI-CAC to non-gated expert segmentations on 102 "Tune-Seg" dataset studies. (**B**) Comparison of non-gated expert CAC annotations to paired gated report CAC scores (combination of "Train-Seg" and "Tune-Seg" dataset patients that had paired gated CT report scores). (**C**) AI-CAC on non-gated vs. paired gated CAC report scores on "Tune-Paired" dataset or (**D**) "Test-Paired" dataset. The AI model was trained only on the "Train-Seg" dataset and hyperparameters were tuned using the "Tune-Seg" and "Tune-Paired" datasets.

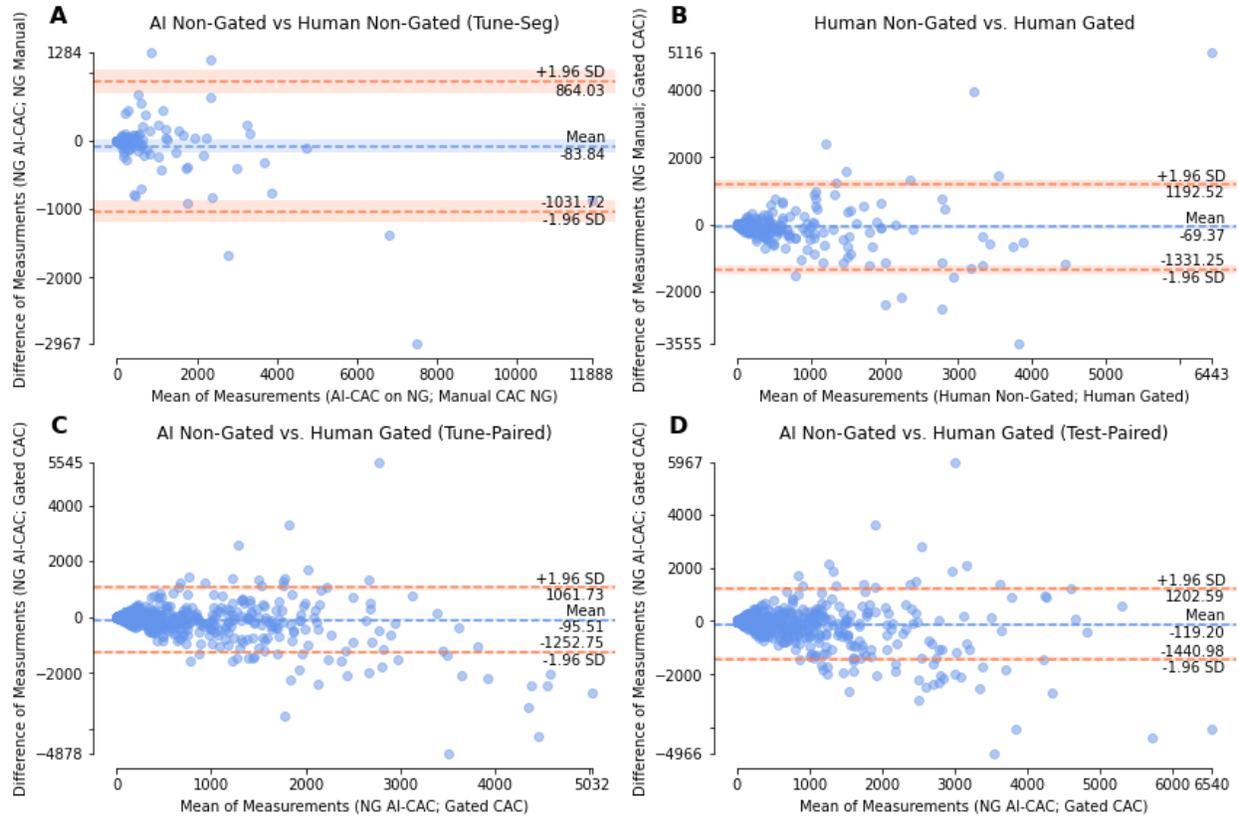

**Supplemental Figure 4. Bland-Altman Plots.**

| CAC Threshold | Reference above Threshold (# Patients) | AI-CAC above Threshold (# Patients) | Accuracy(%) | PPV(%) | NPV(%) | Sensitivity(%) | Specifity(%) | F1(%) |
|---|---|---|---|---|---|---|---|---|
| Toshiba_1 | 290 | 289 | 89.5 | 93.4 | 76.2 | 93.1 | 77.1 | 93.3 |
| Toshiba_100 | 222 | 211 | 86.3 | 90.5 | 80.9 | 86.0 | 86.8 | 88.2 |
| Toshiba_400 | 135 | 136 | 87.9 | 83.1 | 90.7 | 83.7 | 90.3 | 83.4 |
| Philips_1 | 67 | 66 | 93.8 | 97.0 | 78.6 | 95.5 | 84.6 | 96.2 |
| Philips_100 | 51 | 44 | 86.2 | 95.5 | 75.0 | 82.4 | 93.1 | 88.4 |
| Philips_400 | 35 | 31 | 87.5 | 90.3 | 85.7 | 80.0 | 93.3 | 84.8 |
| Seimens_1 | 55 | 49 | 89.0 | 98.0 | 70.8 | 87.3 | 94.4 | 92.3 |
| Seimens_100 | 41 | 32 | 87.7 | 100.0 | 78.0 | 78.0 | 100.0 | 87.7 |
| Seimens_400 | 29 | 14 | 79.5 | 100.0 | 74.6 | 48.3 | 100.0 | 65.1 |
| GE_1 | 227 | 215 | 88.1 | 95.3 | 59.3 | 90.3 | 76.2 | 92.8 |
| GE_100 | 168 | 168 | 88.8 | 91.1 | 85.1 | 91.1 | 85.1 | 91.1 |
| GE_400 | 114 | 108 | 88.8 | 88.9 | 88.8 | 84.2 | 92.3 | 86.5 |
| Men_1 | 618 | 596 | 90.4 | 95.8 | 69.3 | 92.4 | 80.9 | 94.1 |
| Men_100 | 474 | 445 | 87.0 | 92.4 | 79.3 | 86.7 | 87.6 | 89.4 |
| Men_400 | 310 | 285 | 86.8 | 87.0 | 86.6 | 80.0 | 91.6 | 83.4 |
| Women_1 | 21 | 23 | 73.9 | 69.6 | 78.3 | 76.2 | 72.0 | 72.7 |
| Women_100 | 8 | 10 | 91.3 | 70.0 | 97.2 | 87.5 | 92.1 | 77.8 |
| Women_400 | 3 | 4 | 97.8 | 75.0 | 100.0 | 100.0 | 97.7 | 85.7 |
| KVP 120_1 | 574 | 559 | 89.5 | 94.6 | 70.4 | 92.2 | 78.1 | 93.4 |
| KVP 120_100 | 436 | 414 | 86.8 | 91.3 | 80.5 | 86.7 | 86.9 | 88.9 |
| KVP 120_400 | 283 | 269 | 87.9 | 86.6 | 88.7 | 82.3 | 91.6 | 84.4 |
| Non-KVP 120_1 | 65 | 60 | 89.3 | 96.7 | 70.8 | 89.2 | 89.5 | 92.8 |
| Non-KVP 120_100 | 46 | 41 | 91.7 | 97.6 | 86.0 | 87.0 | 97.4 | 92.0 |
| Non-KVP 120_400 | 30 | 20 | 83.3 | 90.0 | 81.2 | 60.0 | 96.3 | 72.0 |

**Supplemental Table 3. Subgroup Analysis.** Performance on "Test-Paired" stratified by Manufacturer, Sex, and KVP.

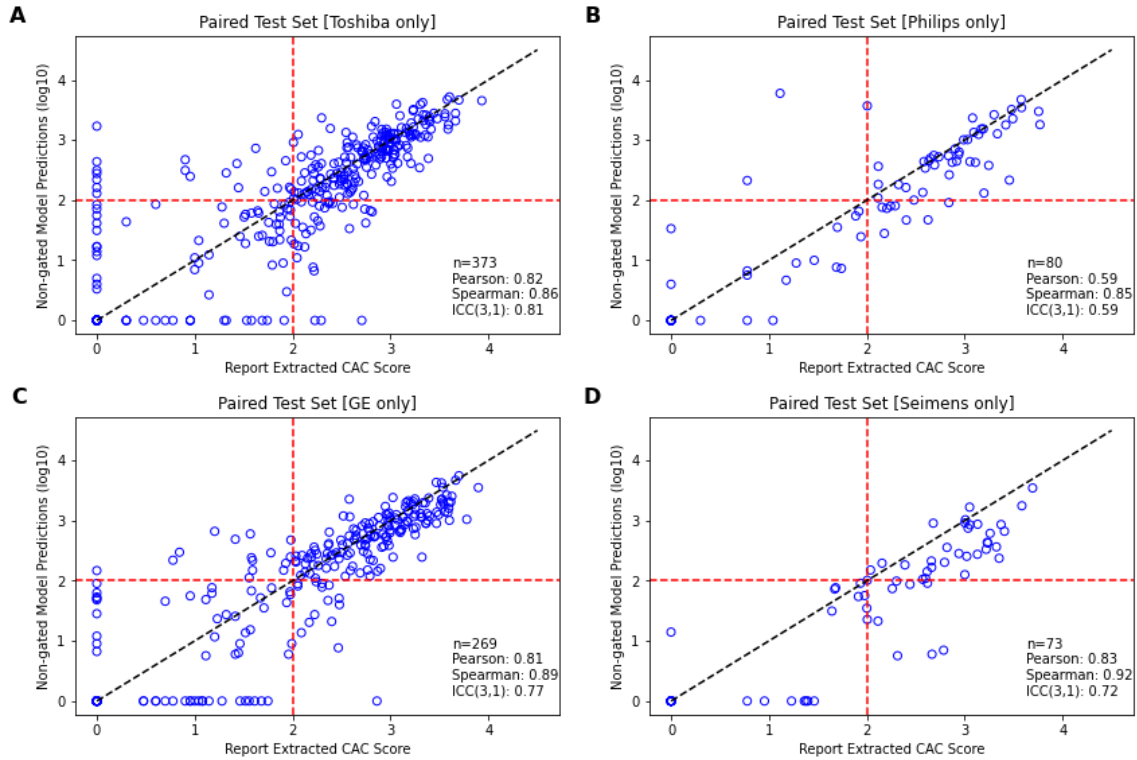

**Supplemental Figure 5.** "Test-Paired" dataset dot-plots separated by the four major CT scanner manufacturers.

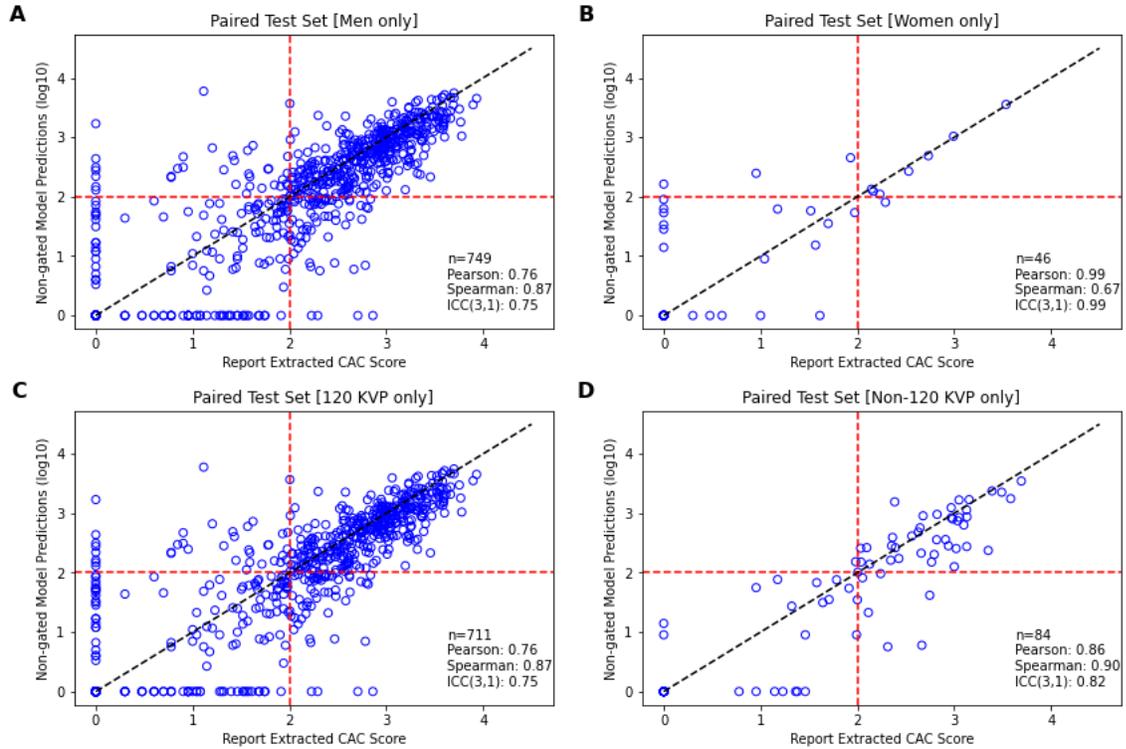

**Supplemental Figure 6.** "Test-Paired" dataset dot-plots separated by Sex **(A, B)**, or KVP 120 vs. Non-120 KVP **(C, D)**.

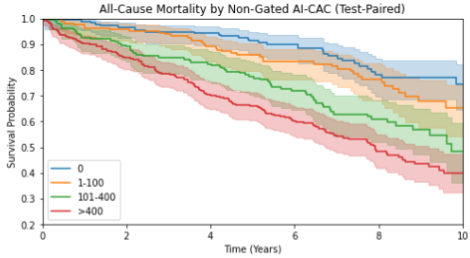
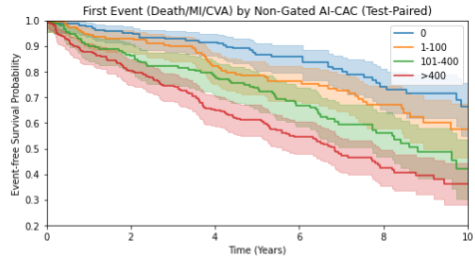
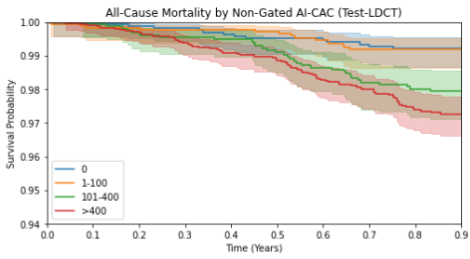
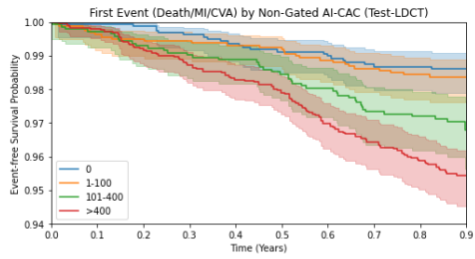
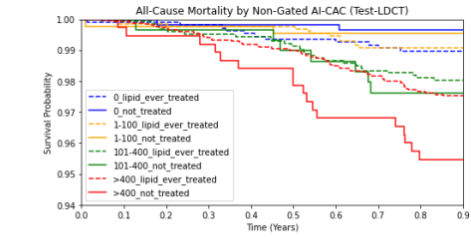
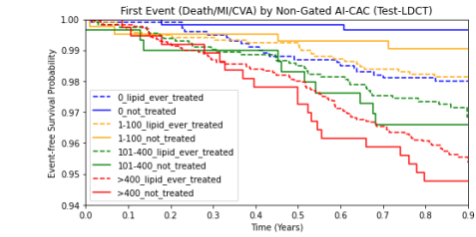

**Supplemental Figure 7**. Univariate KM plots with number at risk counts.

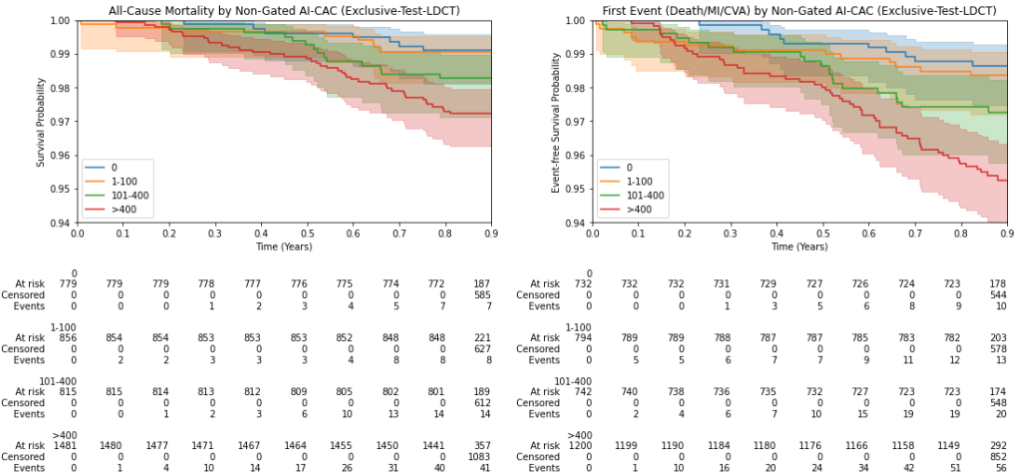

**Supplemental Figure 7 (continued).** Univariate KM plots with number at risk counts for the "Exclusive-Test-LDCT" dataset.

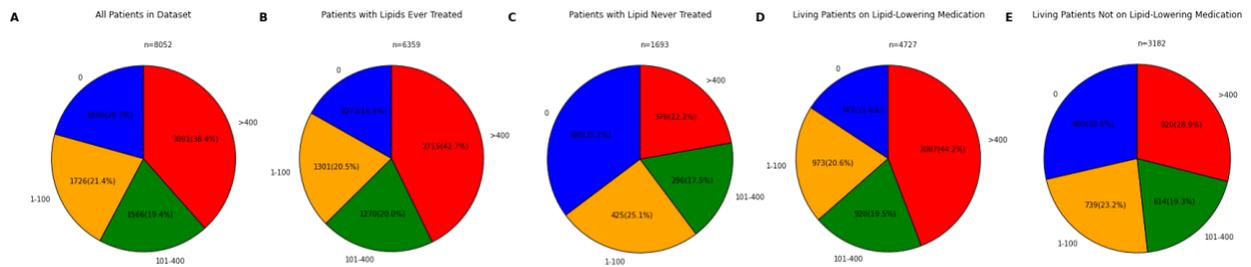

**Supplemental Figure 8. Test-LDCT Dataset AI-CAC Proportions.** **(A)** AI-CAC group breakdown for full dataset, **(B)** breakdown for patients in "Test-LDCT" who ever had a lipid-lowering therapy prescribed, **(C)** never had a lipid-lowering therapy prescribed, **(D)** living "Test-LDCT" patients who were currently on a lipid-lowering therapy, **(E)** living "Test-LDCT" patients currently not on a lipid-lowering therapy.

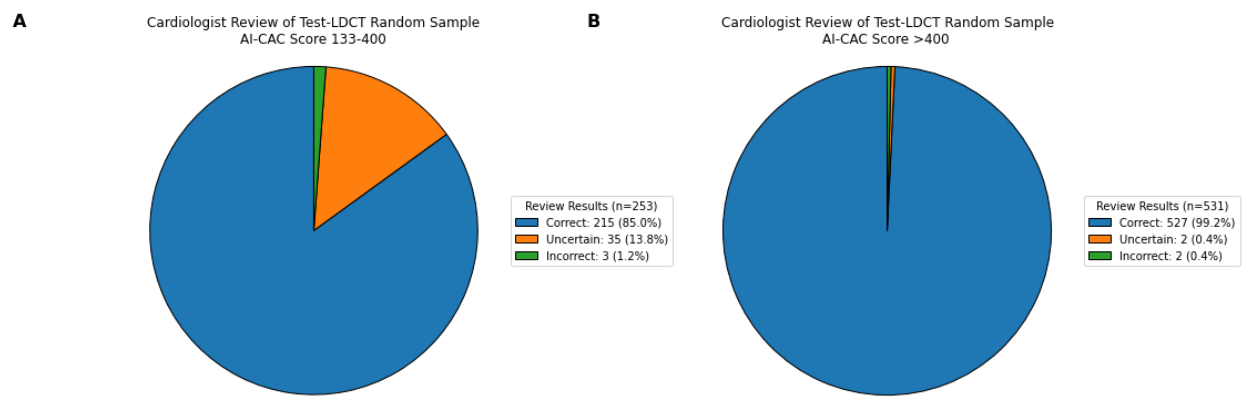

**Supplemental Figure 9. Cardiologist Review of Test-LDCT Dataset Model Predictions.** A total of 784 patient scan predictions were reviewed by four cardiologists individually **(A)** for AI-CAC 133-400 **(B)** AI-CAC >400. "Correct" implies the cardiologist reviewing model predicted masks thought the amount of CAC present on the images was qualitatively sufficient to initiate lipid-lowering therapies, "uncertain" meant further review of images/clinical history would be required, and "incorrect" implied insufficient CAC on shown images to recommend initiation of lipid-lowering therapy.

*Representative examples of true positive and false positive AI-CAC segmentations:*

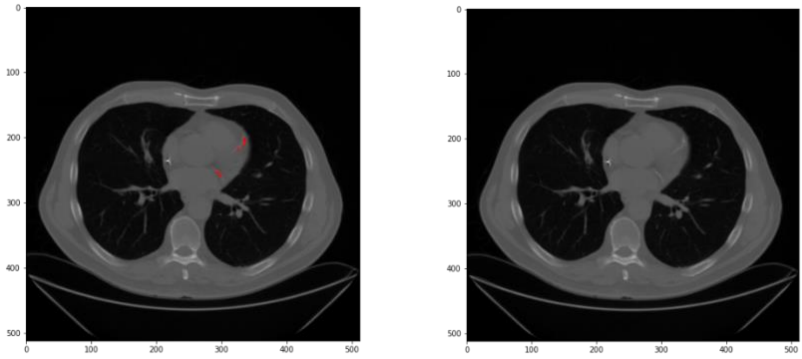

**Example 1**. Model avoids pacemaker wire. Model predicted CAC is shown in red on the left panel.

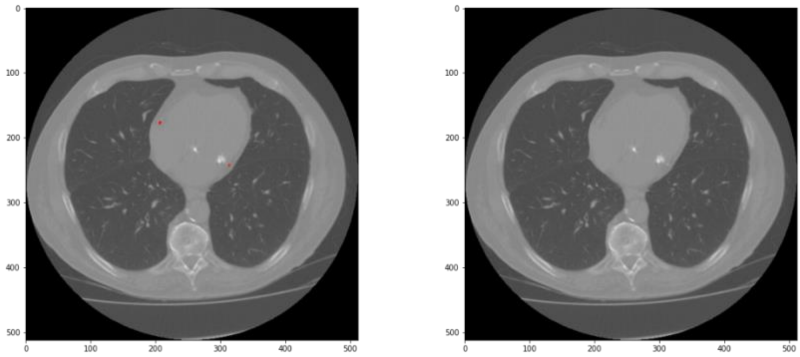

**Example 2**. Model avoids mitral annular calcification.

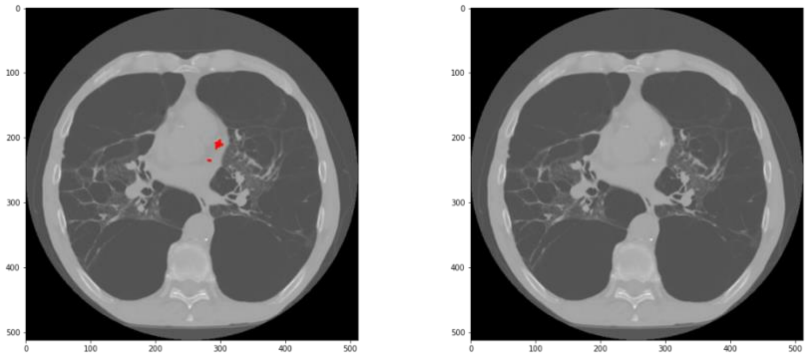

**Example 3.** Model detects CAC despite significant adjacent pulmonary emphysema.

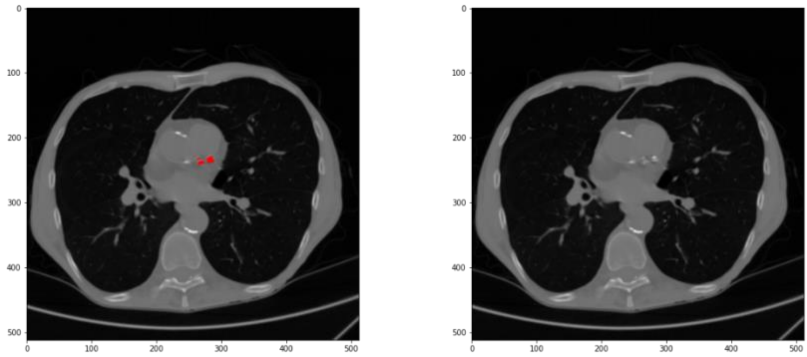

**Example 4.** Model avoids aortic calcium and accurately detects adjacent CAC.

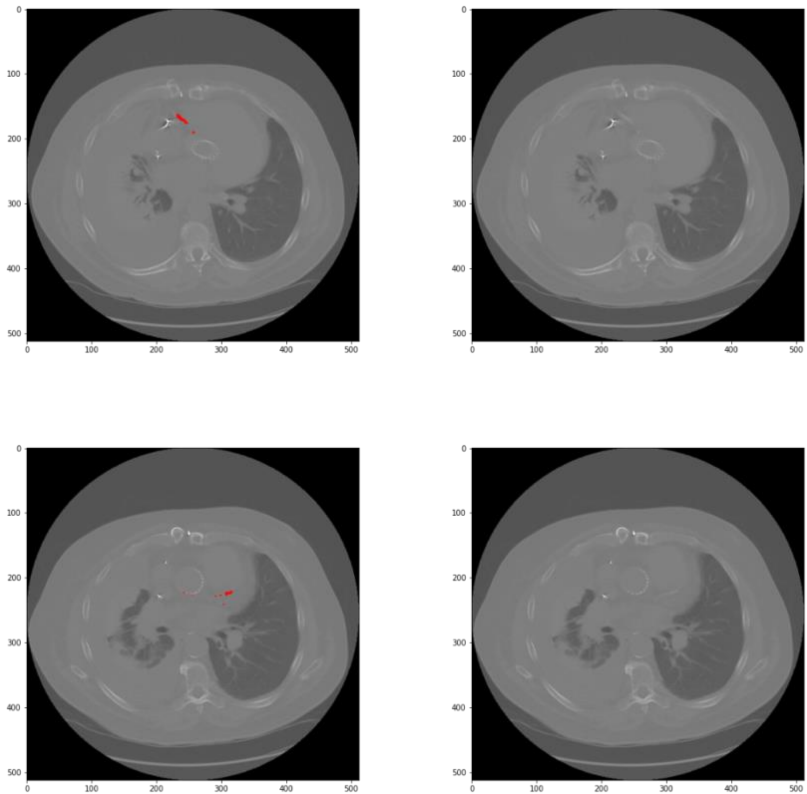

s

**Example 5.** Model predominantly avoids TAVR and pacemaker wires, while accurately segmenting CAC. There is minimal false positive masking of the TAVR in the lower left image panel.

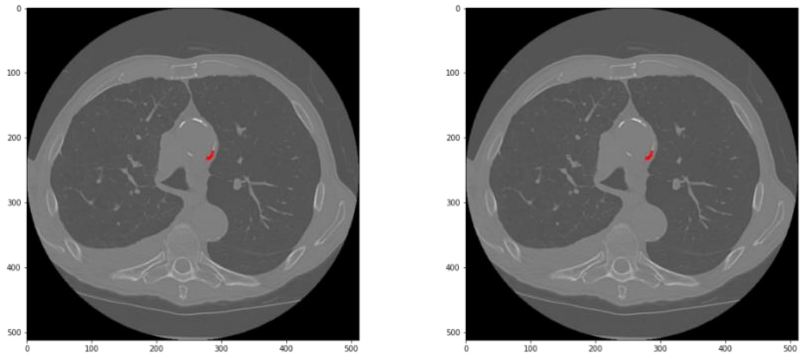

**Example 6.** False positive aortic calcium labeled as CAC. This false positive result is similar in appearance to the angle and thickness of left anterior descending coronary calcium.

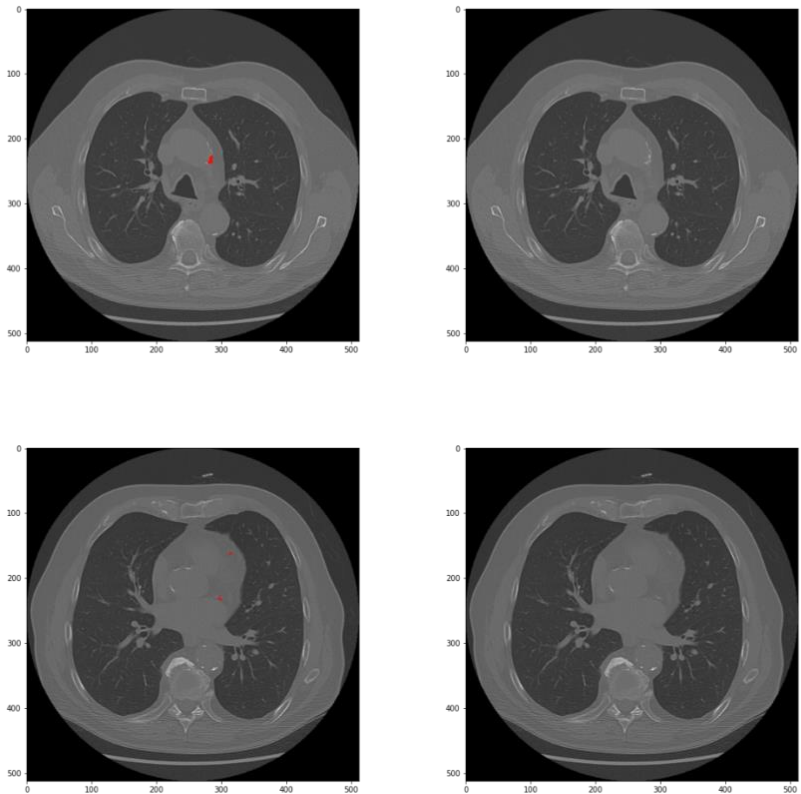

**Example 7**. False positive aortic arch calcium labeled as CAC (top images). Within the same scan, the model correctly avoids ascending and descending aortic calcium (bottom images).